\documentclass{article}

\usepackage{arxiv}

\usepackage[utf8]{inputenc} 
\usepackage[T1]{fontenc}    
\usepackage{hyperref}       
\usepackage{url}            
\usepackage{booktabs}       
\usepackage{amsfonts}       
\usepackage{nicefrac}       
\usepackage{microtype}      
\usepackage{lipsum}
\usepackage{graphicx}
\graphicspath{ {./images/} }

\title{HorGait: A Hybrid Model for Accurate Gait Recognition in LiDAR Point Cloud Planar Projections}

\author{
  Jiaxing Hao $^{1,\ddagger}$, Yanxi Wang $^{2,\ddagger,}$* , Zhigang Chang $^{3}$, Hongmin Gao$^{2}$, Zihao Cheng$^{2}$, Chen Wu$^{2}$ , \\
  \textbf{Xin Zhao$^{4}$ ,Peiye Fang$^{2}$ and Rachmat Muwardi$^{5}$}\\
  $^{1}$ \quad School of Electrical Engineering and Electronics, Shijiahzuang Tiedao University, China;\\
$^{2}$ \quad School of Integrated Circurts and Electronics, Beijing Institute of Technology, China;\\
$^{3}$ \quad Shanghai Jiao Tong University, China;\\
$^{4}$ \quad School of Computer and Information Engineering, Henan University of Economics and Law, China;\\
$^{5}$ \quad Department of Electrical Engineering, Universitas Mercu Buana, Indonesia;\\
$^{*}$ \quad Correspondence: 3220221536@bit.edu.cn; \\
$^{\ddagger}$\quad These authors contributed equally to this work.
}

\begin{document}
\maketitle
\begin{abstract}
Gait recognition is a remote biometric technology that utilizes the dynamic characteristics of human movement to identify individuals even under various extreme lighting conditions. Due to the limitation in spatial perception capability inherent in 2D gait representations, LiDAR can directly capture 3D gait features and represent them as point clouds, reducing environmental and lighting interference in recognition while significantly advancing privacy protection. For complex 3D representations, shallow networks fail to achieve accurate recognition, making vision Transformers the foremost prevalent method. However, the prevalence of dumb patches has limited the widespread use of Transformer architecture in gait recognition. This paper proposes a method named HorGait, which utilizes a hybrid model with a Transformer architecture for gait recognition on the planar projection of 3D point clouds from LiDAR. Specifically, it employs a hybrid model structure called LHM Block to achieve input adaptation, long-range, and high-order spatial interaction of the Transformer architecture. Additionally, it uses large convolutional kernel CNNs to segment the input representation, replacing attention windows to reduce dumb patches. We conducted extensive experiments, and the results show that HorGait achieves state-of-the-art performance among Transformer architecture methods on the SUSTech1K dataset, verifying that the hybrid model can complete the full Transformer process and perform better in point cloud planar projection. The outstanding performance of HorGait offers new insights for the future application of the Transformer architecture in gait recognition.  
\end{abstract}


\section{Introduction}
Gait, as a biometric trait, has garnered considerable research attention for the purpose of individual identification based on their unique walking postures~\cite{chang4}. 
Its distinctive advantages include the capability to be captured remotely through non-intrusive means and its resilience to deliberate disguise, in contrast to other biometric modalities like facial recognition, handwriting analysis, and fingerprint scanning~\cite{chang1,chang2}. 
These qualities highlight the suitability of gait recognition for a range of security applications, including suspect tracking and criminal investigations~\cite{deep}.
In computer vision, gait recognition aims to learn distinctive and invariant representations from the dynamic characteristics of body motion over time. 
Thus, it is essential to first address the fundamental issue of determining the optimal form of input that encompasses all necessary features before applying deep models to learn gait features. 
It is crucial to preserve gait-related characteristics such as body shape, structure, and dynamics while minimizing the influence of gait-irrelevant factors, such as clothing, distance, and perspective, which exist in reality and interfere with recognition~\cite{Authors4,Lidar3D3}.
Previously, research on gait recognition primarily focused on contours, which many researchers considered a 'simple' situation that could be handled by shallow gait models \cite{deep}.
With the advent of gait recognition in the wild, researchers have increasingly recognized that recognition based on shallow 2D features is insufficient for many real-world scenarios, shifting their focus toward deep 3D representation recognition, exemplified by CNN and Transformer architectures. 

\begin{figure}[]
\centering
\includegraphics[width=9.5 cm]{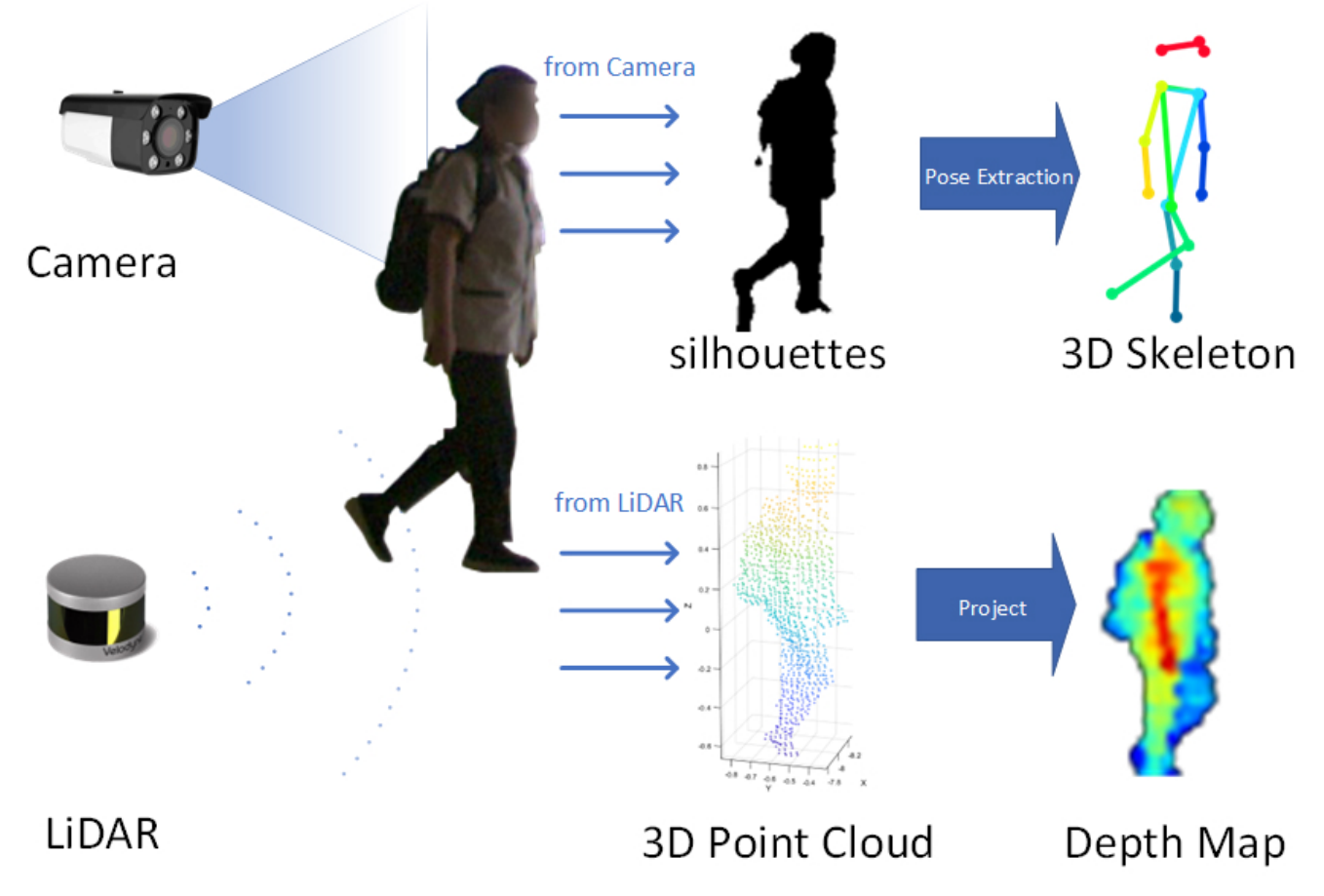}
\caption{The main representations of gait include silhouettes obtained through cameras, manually extracted 3D bones, and 3D point clouds obtained directly by using LiDAR. LiDAR allows for the preservation of the target's original 3D features while also protecting privacy. }\label{fig1}
\end{figure}

The most commonly utilized representation in gait recognition today is the 2D silhouette captured by a camera, which conveys gait using minimal information\cite{Authors7,Authors8}.
While silhouettes have great discriminative capacity due to their explicit preservation of appearance information~\cite{Authors1,Authors2}, they are often vulnerable to interference from clothing and carried items, which can obscure these features. 
Additionally, contour-based methods undoubtedly lose a significant amount of 3D information, such as perspective, motion amplitude, and travel direction, all of which are integral to 3D features. 
As shown in Figure~\ref{fig1}, researchers have proposed various methods that combine contour and 3D representations to address this limitation, including skeletons~\cite{Authors9,Authors10}, 3D meshes~\cite{3D-shape1}, and 3D points~\cite{3D-point1}.
Compared to the other two forms of 3D representation, 3D point clouds can be directly obtained through LiDAR.
LiDAR sensors provide precise 3D body structure information and accurate human perception for gait recognition, while also overcoming the visual blur caused by insufficient lighting and complex backgrounds in outdoor environments, unlike cameras~\cite{lidar1,lidar2}. More importantly, LiDAR does not capture the appearance features of the subject, thereby effectively protecting privacy.

Currently, using deep networks for gait recognition is the mainstream research, and CNN architecture is the most popular method~\cite{deep,deep2,opengait}.
However, CNN-based approaches face challenges in effectively integrating prior information, such as view angle, without introducing extraneous features into the final gait representation. 
Transformer methods, being among the most advanced deep learning approaches, not only address this limitation but also effectively model long-term temporal relationships, which is advantageous for gait recognition in challenging and complex scenarios, such as when certain posture features are occluded or missing~\cite{transformer1,chang3}. 
Currently, only a limited amount of work on contour-based gait recognition has employed the transformer approach, and the results achieved thus far have been unsatisfactory~\cite{deep}. 
This is due to the dummy patch in gait recognition, which severely interferes with the self attention mechanism of traditional Transformer structures.
The fusion of visual Transformer frameworks with CNN to create a hybrid model represents a novel approach to addressing various visual recognition challenges, and utilizing CNN's convolutional segmentation method can help reduce the proportion of redundant patches\cite{Hybrid1,Hybrid2}.

To address the limitations of CNN-based deep recognition methods, we have developed a hybrid approach named HorGait incorporating a Transformer framework for gait recognition using LiDAR. 
For the 3D point cloud data from LiDAR, HorGait employs planar projection to establish connections between points, enabling the Transformer framework to operate effectively. 
In HorGait, we utilized the $g^{\Omega}$  block with input adaptation, long-range and high-order spatial interactions, executed through gated convolution and recursive design. 
The primary contributions of this study are outlined as follows: 

\begin{itemize}
    \item We propose a gait recognition method, called HorGait, which utilizes a hybrid model with a Transformer framework to recognize gait from radar 3D point clouds, addressing the limitations of CNN methods in effectively integrating prior information in gait depth recognition. Compared to other gait recognition approaches, the overall accuracy of HorGait surpasses the existing SOTA Transformer methods.
\end{itemize}
\begin{itemize}
    \item The LHM  block is introduced for input adaptation, long-range, and high-order spatial interaction, enabling HorGait to complete the full Transformer process.  Furthermore, we analyzed the optimal interaction order for LiDAR gait recognition.
\end{itemize}
\begin{itemize}
    \item We conclude that planar projection is more suited to the performance of Transformer framework after evaluating  approaches for processing radar 3D point clouds in various gait recognition methods.
\end{itemize}

\section{Related Works}
\label{sec:headings}

\subsection{Gait Recognition Methods}
GaitBase utilized ResNet to process gait datasets, pioneering the use of deep neural networks in gait recognition~\cite{opengait}, demonstrating robust performance in both ideal and real-world scenarios. During the same period, numerous gait recognition studies based on CNN were also published~\cite{CNN2,CNN3}.
Although deep gait recognition has overcome the limitations of shallow networks, most research still focuses on 2D silhouettes captured by cameras due to their simple form and ease of feature acquisition\cite{Authors7,Authors8}.
This method of recognition, which overlooks the 3D aspects of gait, fails to fully harness the capabilities of deep networks.
Consequently, some researchers have shifted their focus to more comprehensive 3D representations, including meshes\cite{Authors4,3D-mesh1}, point clouds\cite{3D-point1,3D-point4}, and skeletons\cite{Skeletons1,Skeletons2}.
Skeletons and 3D meshes offer an appearance-independent representation that is inherently resilient to motion variations,but they are typically derived from camera images and are influenced by environmental conditions and subjective interpretation, rather than reflecting the original 3D feature. 
In earlier studies, numerous works have employed 3D point sets obtained from LiDAR for gait recognition~\cite{lidar2,Lidar3D2,Lidar3D3}. 
This approach enables the direct capture of 3D gait features, unaffected by environmental conditions or lighting, while preserving the privacy of subjects. 
These advantages have made LiDAR-based gait recognition a prominent focus of current research. 

\subsection{LiDAR-based Applications}
Due to its capacity to capture precise depth information in expansive scenes without being affected by lighting conditions, LiDAR is often used as the primary sensor in autonomous driving and robotics\cite{freegait,Liapp1,lidar2}.
LiDAR plays an important role in 3D perception, and many methods apply it to detection and segmentation\cite{lidar3,Liapp4,Liapp5}.
Gait recognition in the wild can fully utilize the precise depth perception capability of LiDAR, and gait recognition methods can also be installed in a variety of LiDAR devices in the future.
The use of 3D point clouds as input for gait recognition has emerged as a key focus in LiDAR-based gait recognition research in recent years.
Many researchers hold differing views on the input format for recognizing 3D point clouds obtained via LiDAR. 
While some favor directly recognizing 3D point clouds\cite{Lidar3D2,Lidar3D3}, the accuracy of such methods often lags behind contour-based approaches due to an overemphasis on 3D representation, which overlooks the intrinsic connections between points. 
The 3D point cloud can also be projected into the depth image of the LiDAR range view, allowing gait features with 3D structural information to be extracted from the depth image using 2D networks \cite{lidar1}. 
However, \cite{freegait} pointed out that planar projection contains too few dynamic 3D features, leading to the decision to use a multimodal approach combining projection and 3D point cloud data for gait recognition to improve accuracy. 
But multimodal methods require multiple inputs, making them too complex and increasing the cost of practical applications.
The Transformer framework leverages the attention mechanism to capture long-range dependencies within the data, effectively directing the network's focus to the limited 3D features. 
This approach provides a potent solution to the problem of missing unimodal 3D features. 

\subsection{Transformer Framework}
The Transformer framework \cite{transformer4} was originally developed for natural language processing tasks. 
The pioneering work of ViT innovatively applies a Transformer architecture directly to non-overlapping medium-sized image patches for image classification\cite{transformer2}.
The introduction of Swin Transformer signifies the maturity of employing Transformers for computer vision tasks, as it can adapt to the scale of visual elements and perform dense predictions at the pixel level~\cite{swin}. 
Some studies use ViT and Swin Transformer for gait recognition based on silhouettes, achieving good performance on datasets collected from wild environments. However, they tend to perform relatively poorly on constrained gait datasets~\cite{deep,vit1,vit2}. 
While some studies on contour-based gait recognition have incorporated transformer methods, there has been little exploration of using transformers in the analysis of LiDAR depth maps. 
SwinGait introduced a method that combines CNN stages with Transformer stages, yet this approach compromises the completeness of the Transformer process\cite{deep}.
HorNet proposed a hybrid model that employs convolutional network segmentation and high-order interaction to implement the Transformer framework, mitigating the interference of dumb patches while preserving the attention mechanism's integrity\cite{Hor}. 
This hybrid model that blending CNN and Transformer framework, offers fresh insights for LiDAR-based gait recognition. 

\section{Materials and Methods}
\label{sec:others}
\subsection{Dataset Description}
SUSTech1K is the first large-scale LiDAR-based gait recognition dataset, collected using a LiDAR sensor and an RGB camera across three real-world scenes\cite{lidar1}. The dataset contains 25,239 sequences with diverse variations such as visibility, viewpoints, occlusions, clothing, and carrying items. This extensive variability ensures that models trained on SUSTech1K face a broad range of real-world scenarios, enhancing their robustness and generalizability. The dataset is divided into a training set with 250 subjects and 6,011 sequences, and a test set with 800 subjects and 19,228 sequences, providing comprehensive data for LiDAR depth map gait recognition research.

\subsection{Dataset Preparation}
The 3D point cloud containing gait information, collected by the LiDAR, consists of a set of points with (x, y, z) coordinate information\cite{lidar2,Lidar3D2,Lidar3D3}. 
While some works opt to use this point cloud directly for recognition, the 3D point cloud inherently lacks coherence, making it impossible to directly connect the points representing a specific body movement from one frame to those in the next. 
As a result, even though these methods strive to emphasize the 3D features of gait, achieving high accuracy in identification remains challenging. 
Currently, the most effective method for recognizing long-term temporal associations is still centered in the two-dimensional domain. 
The key issue in LiDAR-based gait recognition research is finding a way to retain 3D features to the greatest extent while ensuring the connectivity between points. 
In previous studies, the most widely used point cloud dimensionality reduction method was plane projection\cite{lidar1}. 
This method compresses the point cloud horizontally, using the distance from each point to the radar as the color, effectively establishing the connection between the point cloud and gait while preserving 3D features. 
The 3D point cloud used for gait recognition in the dataset is obtained from the VLS128 LiDAR scanner and consists of a collection of points with (x, y, z) 3D coordinate information.
The point cloud set can be expressed as $P=\{P^j_i|i=1,2,...,M;j=1,2,...,n_i\}$,where $M$ is the number of identities and $n_i$ is the number of sequence of each $i$-th identity.
Each point cloud sequence$P^j_i\in \mathbb{R}^{T*N*C}$,is with $T$ frames and $N$ points for each frame,where $C$ represents the number of feature channels.
For the coordinates of a 3D point  $p=(x.y,z)^T$ in the dataset $P$.
Since LiDAR uses a cylindrical coordinate system to collect point sets, the horizontal and vertical coordinates of the corresponding depth map in the plane projection are as follows:
\begin{equation}
\begin{split}
    h=arctan(x,y)/\Delta\theta   \\
    v=arcsin(z,R)/\Delta\phi
\label{eq:3-1} 
\end{split}
\end{equation}
The coordinates $(h,v)$ are the positions of the corresponding two-dimensional pixels in the depth map mapped to the point $p$ based on spherical projection.
The $\Delta\theta$ and$\Delta\phi$ represent the average resolution of the horizontal and vertical angles between successive beam emitters of LiDAR.
The range $R=\sqrt{x^2+y^2+(z-c)^2}$ is the average distance between the point set and the origin within the range of  LiDAR.
Each element at position $(h,v)$ in the map is filled with an RGB color selected based on $D$,where $D=\sqrt{x^2+y^2}$.
The depth projection is then normalized and transformed from single-channel images into RGB images. 

\subsection{Description of Models}
\begin{figure}[]
\centering
\includegraphics[width=9.5 cm]{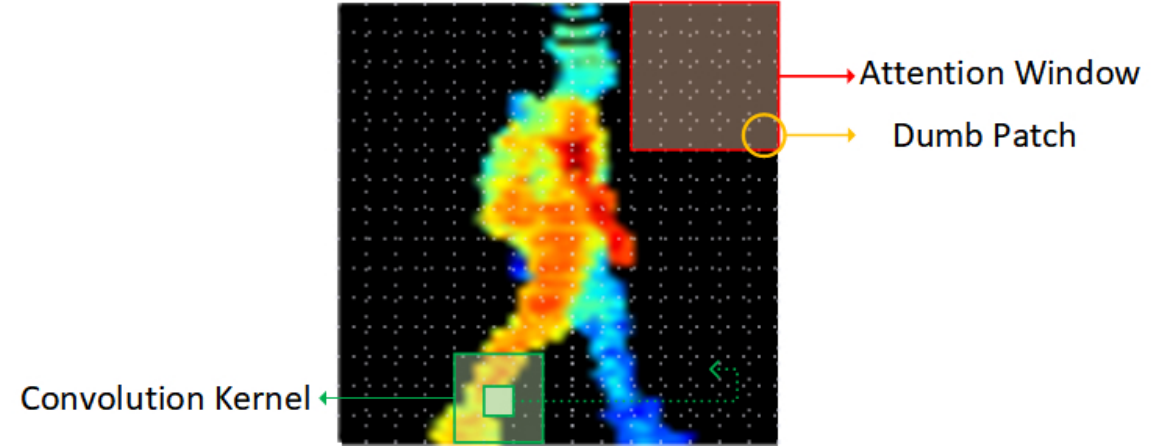}
\caption{Gait recognition datasets often contain a large number of information-free areas, leading to numerous 'dumb patches' in the Transformer architecture. This significantly increases the risk of generating useless or invalid gradients during the self-attention calculation process. The convolution kernel's movement in convolutional networks can help reduce the number of these dumb patches. To address this issue, current solutions include integrating CNNs into the Transformer framework.} \label{fig4}
\end{figure}

With the application of visual transformers, the shift from CNNs to Transformer frameworks has become a major trend in computer vision research \cite{swin, transformer2, transformer3}. 
However, in gait recognition, the presence of 'dumb' patches has limited the performance of Transformers. 
Unlike other tasks using float-encoded RGB images, the 2D input for gait recognition often contains large all-white or all-black regions that may exceed the area containing the desired feature representation. 
Figure~\ref{fig4} illustrates how these uninformative regions create massive 'dumb patches' in the Transformer architecture, significantly increasing the risk of generating useless or even invalid gradients during the self-attention calculation. 
\cite{deep} proposed a method that first passes through two layers of CNN blocks and then uses a Transformer. This approach alleviates the phenomenon of dumb patches to some extent, but the Transformer process is not comprehensive enough to fully highlight its unique advantages. 
HorNet built a hybrid model by using the Recursive Gated Convolutions ($g^n$ Conv) block to replace the self-attention block in the Transformer, achieving better performance than the Swin Transformer \cite{Hor}. 
A key difference between Vision Transformer and previous architectures is that Vision Transformer enables high-order spatial interactions in each basic block.
The $g^n$ Conv leverages this principle to implement arbitrary $n$-order interactions while avoiding the quadratic complexity associated with self-attention. 
Inspired by this methodology, HorGait has conceived LHM(LiDAR Hybrid Module) Block by substituting conventional convolutional layers with advanced deep CNN blocks within the hybrid model. LHM Block effectively mitigates the 'dumb patch' issue through its sliding window mechanism, hierarchical dense feature extraction, and expansion of the receptive field. 

\begin{figure}[]

\centering
\includegraphics[width=16.5 cm]{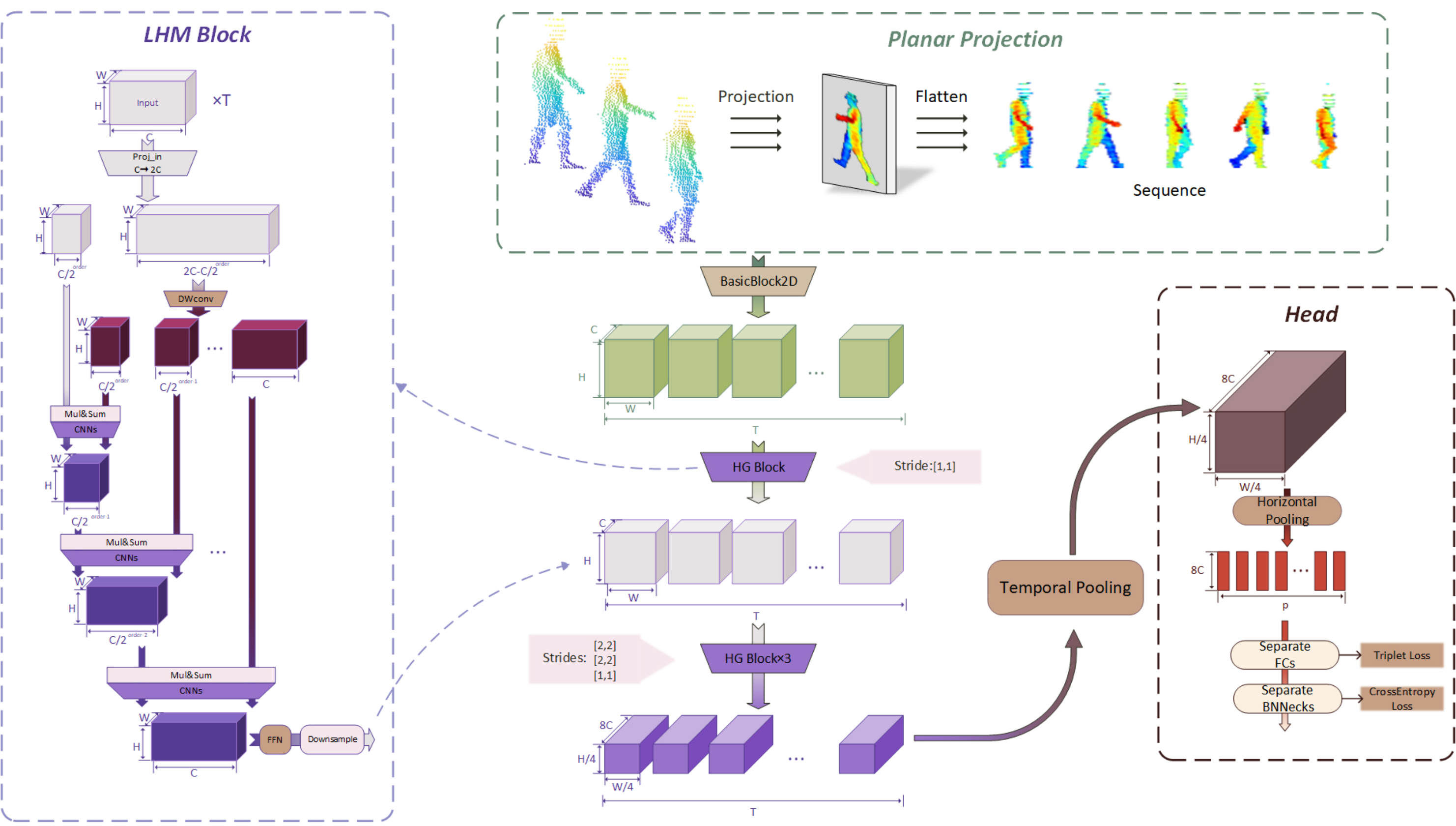}

\caption{The position and radius of the reference sphere influence the spherical projection. The z-axis height of the reference sphere determines the height of the compression center in the projection; too high or too low will result in missing points.  \label{fig5}}
\end{figure}

We present HorGait to process gait sequences derived from LiDAR depth maps utilizing the Transformer architecture and deep CNN blocks. 
As illustrated in Figure~\ref{fig5}, HorGait processes a LiDAR depth map of size 64×44 through a convolutional and normalization network before feeding it into the hybrid model's network layer. 
Each layer comprises several blocks, analogous to the multi-head self-attention modules in the Transformer architecture, employed to model the spatial interactions between visual data and CNNs. 

\subsubsection{Pipline}
The principle of high-order interaction of LHM Block is shown in Figure~\ref{fig5}.
Let$x \in \mathbb{R}^{HW\times C}$   be the input after convolution layer and normalization,  the output of the gated convolution$y=gConv(x)$.
\begin{equation}
\begin{split}
 [p^{HW\times C}_{0},q^{HW×C}_{0}]= \phi _{in}(x)\in \mathbb{R}^{HW\times2C}\\
         p_{1}=f(q_{0})\odot  p_0+p_0\in \mathbb{R}^{HW\times C}\\
         y=\phi _{out}(p_1)\in \mathbb{R}^{HW\times C}
\label{eq:3--1} 
\end{split}
\end{equation}
where $\phi_{in}(x)$, $\phi_{out}(x)$are linear projection layers to perform channel mixing, $f(x)$ is a depth-wise convolution.
Note that $p_1^{i,c}=\sum_{j\in \Omega_i}w^c_{i\rightarrow j}q_0^{i,c}p_0^{i,c}$is the local window centered at $i$ and $w$ represents the convolution weight of $f(x)$.
Hence, the aforementioned formula explicitly incorporates the interaction between adjacent features $p_0(i) $and $q_0(j)$ through element-wise multiplication. 
We regard the interaction in gConv as a 1-order interaction since each $p_0(i)$ has interacted with its neighboring feature $q_0(j)$ only once.

Design higher-order interactions based on the 1-order interactions above, using recursive gating  $g^{\Omega}$ Conv.
$g^{\Omega}$ Conv is a recursive gated convolution block that enhances the model's capacity by integrating higher-order interactions. 
In a formal manner, we initially use $\phi_{in}(x)$ to derive a set of projected features $p_0$ and $\{q_k\}^{\Omega-1}_{k=0}$,$\Omega$ is the numner of orders.
\begin{equation}
\begin{split}
    [p_0^{HW\times C_0},q_{0}^{HW\times C_0},...,q_{n-1}^{HW\times C_{n-1}}]=\phi_{in}(x)\\
    \phi_{in}(x)\in \mathbb{R}^{HW\times (C_0+\sum_{0\geq k \leq n-1}C_k)}
\label{eq:3-2} 
\end{split}
\end{equation}
Then the gated convolution block is performed recursively by
\begin{equation}
\begin{split}
    p_{k+1}=(f(q_k)\odot g_k(p_k)+g_k(p_k))/\alpha,k=0,1,...,\Omega -1
\label{eq:3-3} 
\end{split}
\end{equation}
where we scale the output by $1/\alpha$ ued to stabilize the training. $f(x)$ is a depth-wise convolution layer and $g_k(x)$ is deep CNNs block.
Ultimately, we input the output of the final recursion step $q_n$ into the projection layer $\phi_{out}$, resulting in the outcome of the $g^{\Omega}$ Conv.
From the recursive formula in Equation \ref {eq:3-3} , it is evident that the interaction order of $p_k$ increases by 1 with each step.
The above is the process of $ g^{\Omega}$Conv to achieve $\Omega$-order spatial interaction.
To prevent high-order interactions from introducing excessive computational overhead, we define the channel dimension at each order as follows:
\begin{equation}
\begin{split}
    C_k=\frac{C}{2^{\Omega-k-1}} 
\label{eq:3-4} 
\end{split}
\end{equation}
After completing the $g^{\Omega}$ process, the result is added to the input of LHM Block and enters the next Block. 
Each layer of the network must pass through a downsampling network before output to ensure seamless transmission between input and output. 
\subsubsection{Model Architectures}
\textbf{Depth-wise convolution layer}. 
The deep convolutional layer facilitates large kernel convolution, akin to the local window used for self-attention in Transformers, and is represented as the 'DWcov' layer in Figure~\ref{fig5}.
This local window grants the Transformer a broad receptive field, enabling it to more effectively capture long-term dependencies— one of the primary advantages of visual Transformers.
In this work, a 7×7 Convolution is selected as the depth-wise convolution $f(x)$ of $ g^\Omega$Conv to capture these extended interactions. 
The 7×7 kernel is the default window size for Swin Transformers\cite{swin} and the kernel size for the deep CNN model ConvNext\cite{deep2}. This kernel size has demonstrated strong performance across various studies\cite{Hor,7*7}. To ensure optimal interaction with the CNNs block, this work also adopts this structure. 

\begin{figure}[]
\centering
\includegraphics[width=12.5 cm]{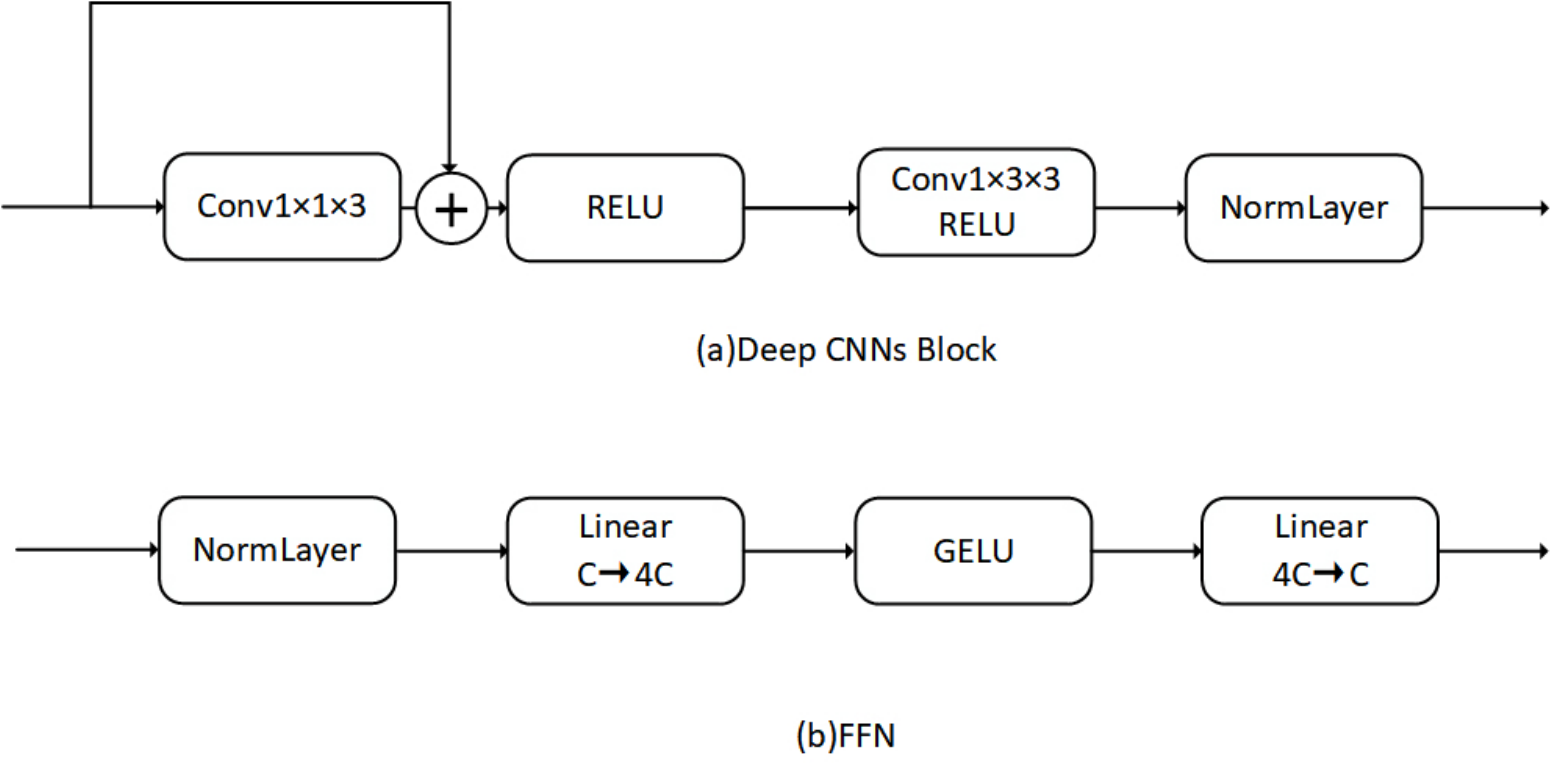}
\caption{(a) The CNN network structure in LHM Block comprises two convolutional networks with varying convolution kernel sizes.(b) The FFN layer in LHM Block consists of two linear layers and a GELU activation layer.} \label{fig6}
\end{figure}

\textbf{Deep CNNs block}.
The essence of CNNs lies in their dense feature extraction and receptive field expansion through a sliding window mechanism and hierarchical architecture, enabling them to effectively address the challenges posed by dumb patches. 
The Deep CNNs block, positioned after the gated convolutional recursive operations, performs dense feature extraction and receptive field expansion for the results of each order of interaction, preventing the loss of valuable information.
As shown in Figure~\ref{fig6}(a), deep CNNs block composed of pseudo 3D residual units, and the pseudo 3D residual unit is composed of two 1D temporal and a 2D spatial convolution layers.
From the recursive formula in Equation \ref {eq:3-3} ,the channels of $p_k \in \mathbb{R}^{HW\times C_{k-1}}$ is different from $f(q_k) \in \mathbb{R}^{HW\times C_{k}}$ because the result of interaction share the size with $f(q_k) $.
Therefore,there is a variation in the size of  $g_k(x)$ ,although they share the same structure:
\begin{equation}
    g_k=CNNs(C_{k-1},C_k),\\k=1,2...,n-1
\label{eq:3-5} 
\end{equation}

\textbf{FFN}.
The Feed-Forward Neural Network (FFN) further performs a non-linear transformation on the representation of each position within the Transformer to enhance the model's expressive capacity.
As illustrated in  Figure~\ref{fig6}(b), the FFN consists of a NormLayer and two linear layers. The first linear layer quadruples the number of channels, applies GELU activation, and then the second linear layer restores the original number of channels.
This FFN structure effectively enhances the model's expressiveness, allowing it to learn more complex features and patterns. 

\section{Experiments and Analysis}

HorGait can be divided into two main components: the hybrid model incorporating the Transformer framework LHM Block and the projection of the LiDAR 3D point cloud . 

\subsection{Hybrid Model Stage}
\begin{figure}[]
\centering
\begin{minipage}[]{1\linewidth}
\centering
\includegraphics[width=16.5 cm]{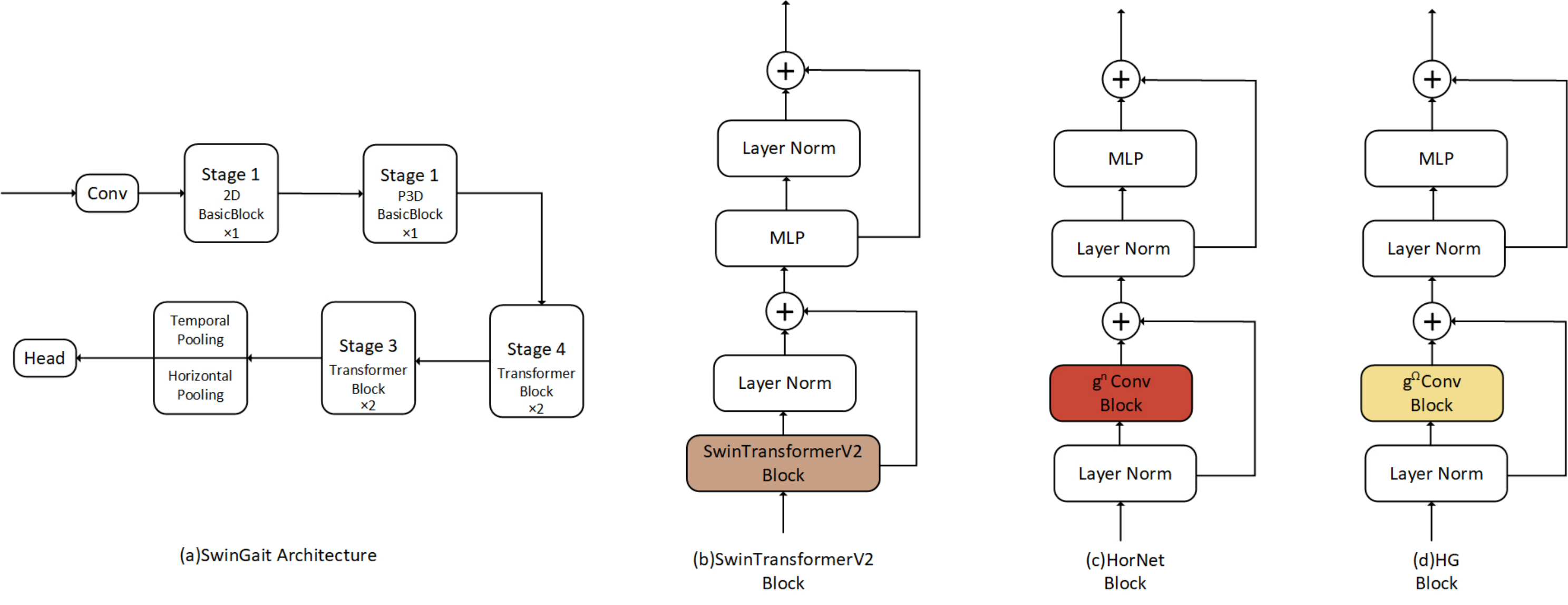}
\end{minipage}

\caption{(a) The network structure of SwinGait, where the backbone consists of four stages. (b) The structure of SwinTransformer V2 block (c) HorNet block (c) LHM block. This work uses these blocks to replace the stages in the SwinGait network structure to construct the Transformer method and the hybrid model method.} \label{fig7}

\end{figure}

\cite{freegait} pioneered the structure of the visual Transformer method for gait recognition, as illustrated in Figure~\ref{fig7}(a). 
In this framework, after the dataset is processed by an initial convolutional network layer, it progresses through four stages: a 2D CNN block, a pseudo 3D CNN block, and two Transformer blocks.
The structure of SwinGait undergoes several layers of CNN stages before passing through the Transformer stage, significantly reducing the number of dummy patches prior to employing the Transformer. 
However, two stages are insufficient for a complete Transformer process. 
Hence, this work replaces all four stages with Transformer blocks or hybrid model blocks incorporating Transformer architecture to evaluate the performance of various methods. 

\begin{figure}[]

\centering
\includegraphics[width=17.5 cm]{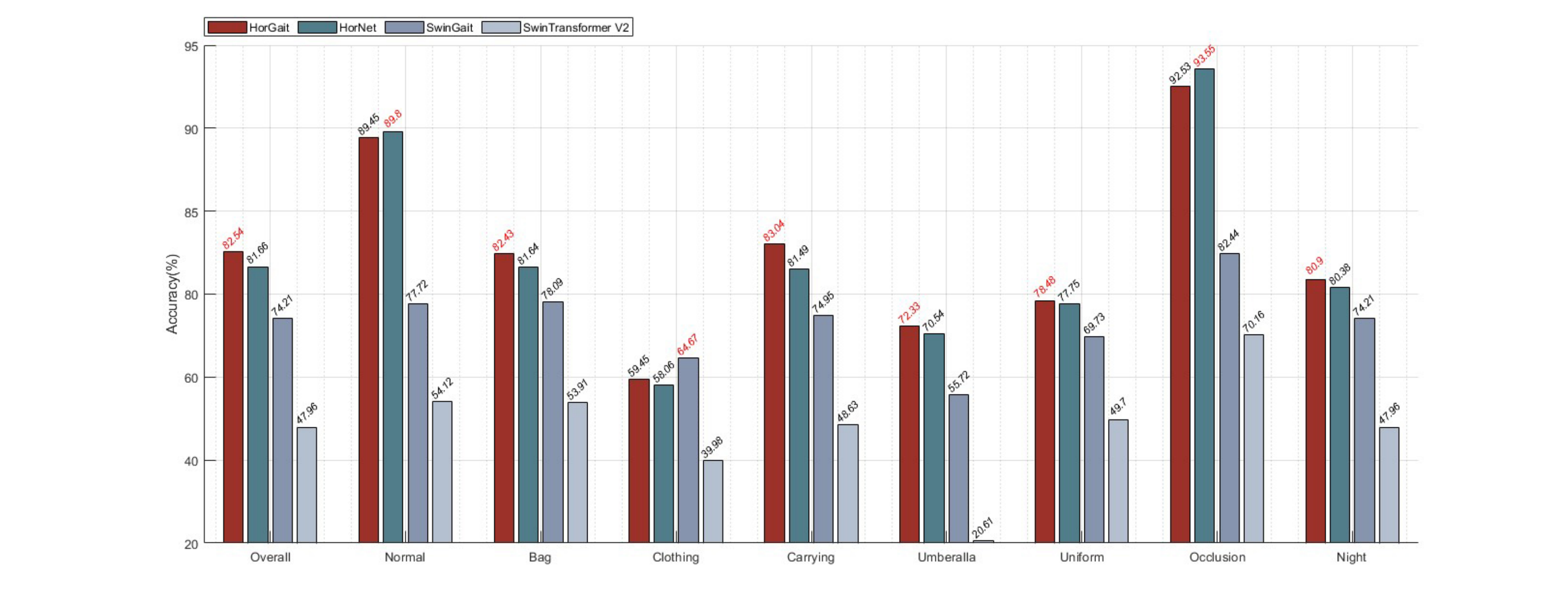}
\caption{HorGait, HorNet, SwinGait, and SwinTransformer V2 compare the accuracy of LiDAR gait recognition in various situations. Except for HorNet achieving better results under Normal conditions and SwinKit achieving better results under Clothing conditions, our work outperforms other methods with Transformer frameworks.} \label{fig8}

\end{figure}

Figure~\ref{fig8} illustrates the performance of SwinGait, Swin Transformer V2, HorNet, and HorGait on the LiDAR-projected depth map\cite{deep,swin2,Hor}. In terms of overall accuracy, HorGait employing LHM Block outperforms the state-of-the-art (SOTA) methods for gait recognition using the Transformer framework. 
Under specific conditions, HorGait achieved the highest accuracy in Bag, Carrying, Umbrella, Occlusion, and Night scenes. 
In particular in the Occlusion scene, where dumb patches are more prevalent due to obstructions, HorGait improved by 11.36\% compared to the existing SOTA Transformer method SwinGait.
This demonstrates that HorGait has made significant progress in overcoming the dumb patch issue.

Compared to methods like SwinGait that employ multiple CNN stages, HorGait features a more comprehensive Transformer framework, further enhancing the effectiveness of high-order interaction of attention.
HorGait employs CNN network layers in the high-order interactions across all four stages, providing a more effective dumb patch elimination than SwinGait, which incorporates CNN blocks only in the first two stages. This advantage is particularly evident in scenarios with more dummy patches, such as Normal and Occlusion, where HorGait improves by over 10\% compared to SwinGait. 
Another major advantage of HorGait over SwinGait is its stronger attention mechanism. In conditions with significant interference, such as Umbrella, where gait occupies a smaller portion of the dataset, only enhanced attention can yield more accurate recognition. HorGait surpasses SwinGait by 16.61\% in the condition, respectively. 
But it is also worth noting that in the clothing scenario, HorGait performs worse than SwinGait, possibly because the attention mechanism represented by high-order interactions may have a counterproductive effect in cases involving subtle changes. 

In contrast to the traditional visual Transformer methods represented by Swin Transformer, the convolution kernel segmentation technique used in HorGait better addresses the dumb patch problem in gait recognition.
By comparison, we can see that the dumb patch has a significant impact on the Transformer. Particularly in the Umbrella scene, where the attention mechanism is crucial, gait features are further compressed by the dumb patch, leading to a recognition accuracy of only 20.61\%. 

Compared to the hybrid models of Transformer frameworks like HorNet, HorGait incorporates a pseudo 3D CNN layer that is more suited for gait recognition on RGB depth maps, and optimizes the interaction algorithm to strengthen its interactive capabilities. 
It can be seen that except for the Normal and Occlusion conditions, the performance of the two is comparable, and HorGait surpasses HorNet under other conditions.
The presence of more dumb patches in these two conditions suggests that HorGait introduces additional interactions to improve recognition accuracy in special cases, but this comes at the cost of reduced resistance to dumb patches. 

\subsection{Order}
The $g^n$Conv block proposed by HorNet can realize Transformer-like interactions of arbitrary orders with bounded complexity\cite{Hor}, so LHM Block developed based on this structure in HorGait also has this advantage. 
Generally, there are two approaches to determine the order: applying a consistent order across all stages or implementing a step-by-step order within the stages. 

\begin{figure}[]
\centering
\includegraphics[width=12.5 cm]{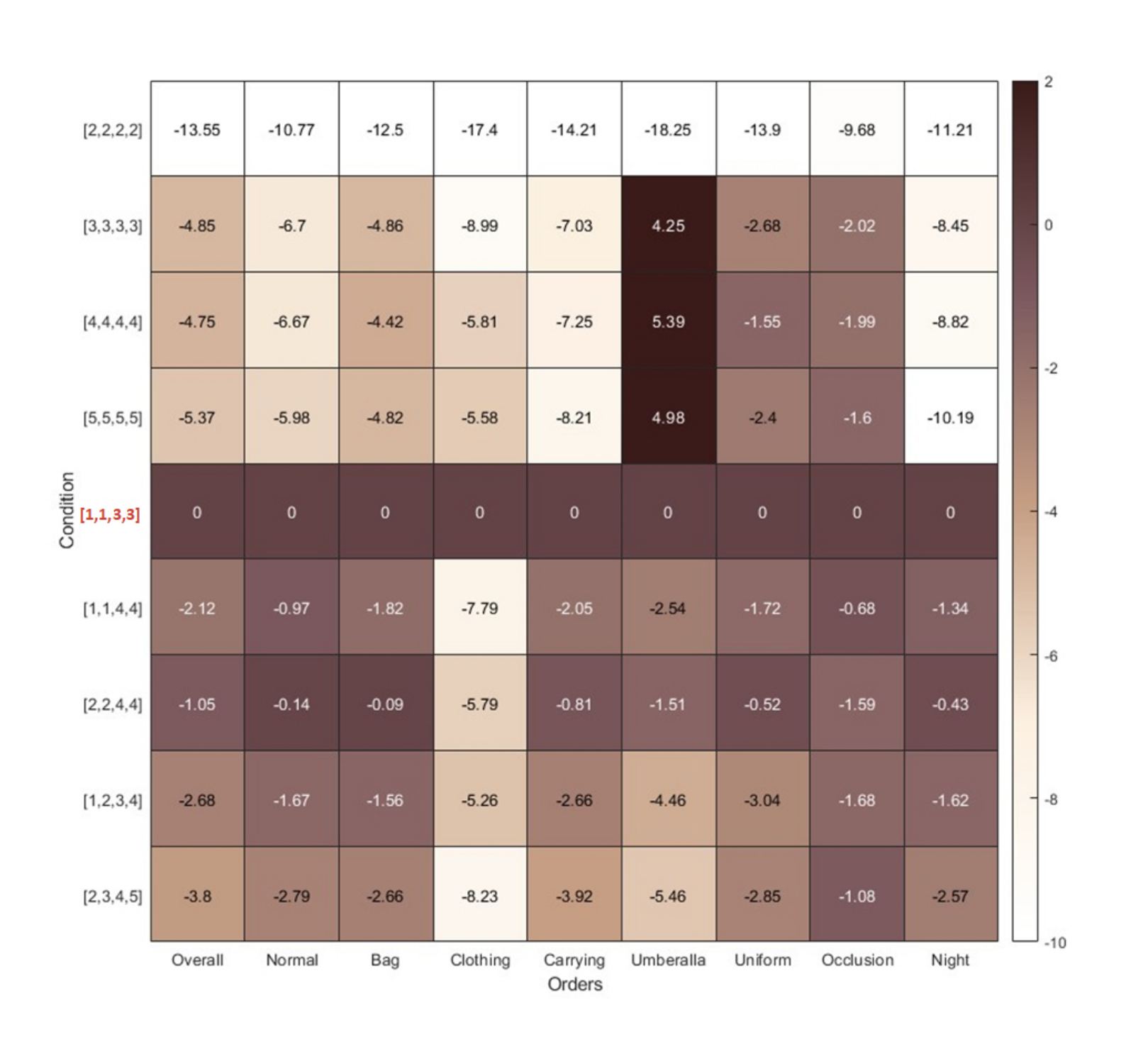}
\caption{The relative recognition accuracy of the eight order combinations in various situations, with the [1,1,3,3] combination achieving the highest overall accuracy as the benchmark. It is evident that the performance of single-order combinations generally follows a step-by-step pattern. The results also indicate that a higher number of orders is not necessarily beneficial when the number of channels is limited.  \label{fig9}}
\end{figure}

As illustrated in Figure~\ref{fig9}, we compare 4 multi-stage models with a uniform order and 5 models with progressive orders, plotting accuracy comparisons across various conditions. 
Compared to the model with a consistent order throughout, the model with progressive orders performs better, with the order [1,1,3,3] achieving the highest overall accuracy. 
This is because the number of channels progressively increases across stages, and maintaining the same interaction order makes it difficult to meet the channel distribution requirements at different stages. 

When each stage utilizes the same order, the optimal result is achieved with an order of 4. 
This suggests that higher-order interactions are not necessarily advantageous for LiDAR gait recognition, likely due to the increased impact of 'dumb patches' in higher-order.
This suggests that employing excessively high interaction orders in layers with fewer channels is unwise, as such configurations can allocate a disproportionate amount of influence to low-impact interactions, thereby diminishing recognition accuracy. 
However, it is also observed that under the umbrella condition, the same-order combination performed better, indicating that the higher-order interaction in the same-order combination can generate stronger attention in specific environments. 

In the progressive order combinations, there are four-step progressions such as [1,2,3,4] and [2,3,4,5], and two-step progressions such as [1,1,3,3], [1,1,4,4], and [2,2,4,4]. 
By comparison, it can be seen that the performance of the four-stage progression is worse than the uniform order combination and not as good as the two-stage progression. 
This is because, during the channel expansion process, the number of channels doubles, causing the lowest-order split size in the four-step progression to remain constant. 
The two-stage progression avoids this issue, allowing different low-order splits to introduce variability and improve the robustness of network recognition. 
Finally, we compared three two-step progressive order combinations and found that [1,1,3,3] enables HorGait to achieve optimal performance in LiDAR gait recognition. 

\subsection{Projection}
\begin{figure}[]
\centering
\includegraphics[width=9.5 cm]{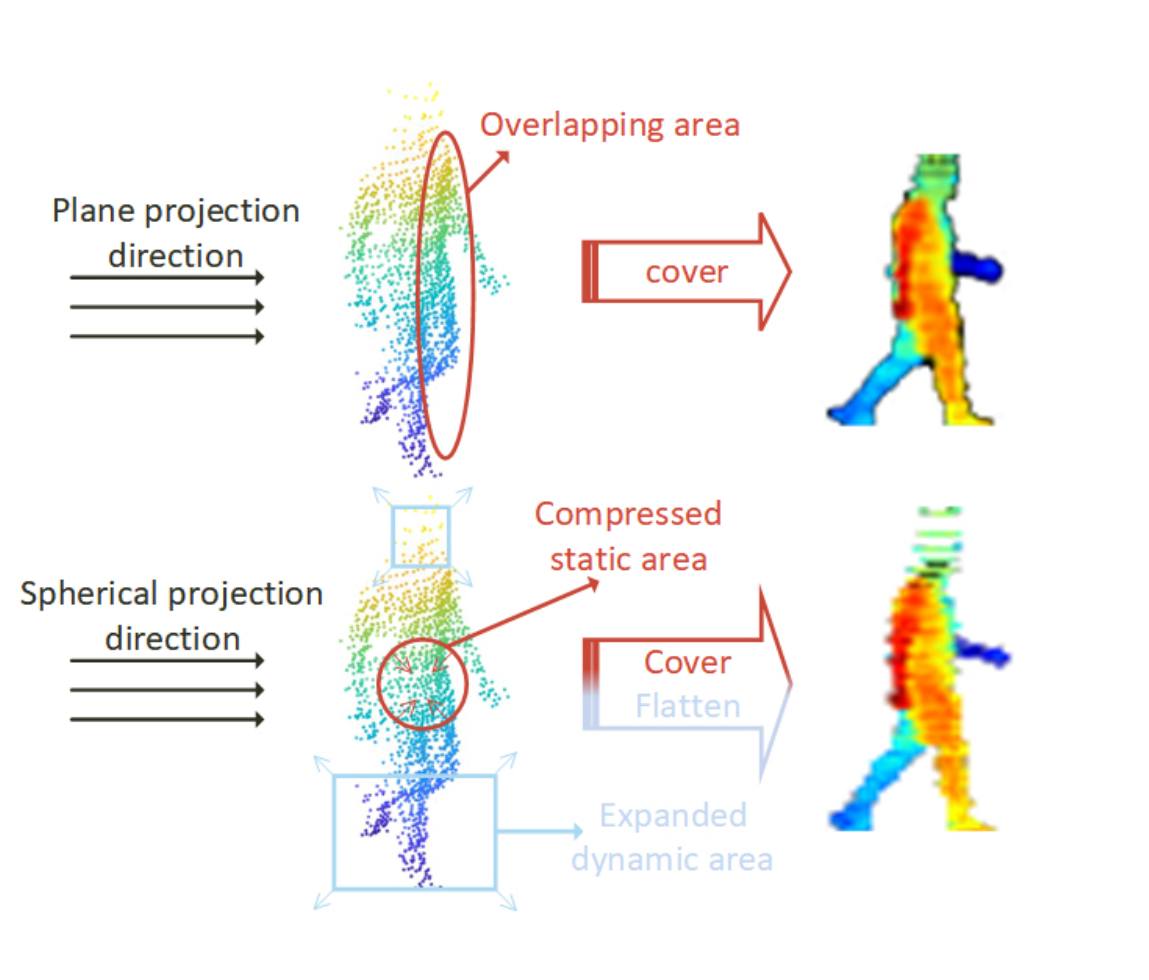}
\caption{Comparison between planar projection and spherical projection: horizontal plane projection directly covers overlapping points, leading to the loss of 3D dynamic features. In contrast, spherical projection, due to its non-uniform projection, compresses static areas and expands the point set of dynamic areas, avoiding the loss of points containing 3D dynamic features. \label{fig2}}
\end{figure}

Since point clouds inevitably overlap horizontally in a planar projection, the resulting projection will obscure some overlapping points, leading to a loss of 3D features\cite{freegait}. As illustrated in Figure~\ref{fig2}, our previous work introduced a non-planar projection method that can expand certain areas while compressing others\cite{spherigait}.
This approach enables the coverage of static regions while dispersing dynamic areas, thereby preserving a greater amount of 3D features. 
This projection method proves highly effective in CNN networks, surpassing planar projection in performance across numerous gait recognition models. 

Building on formula \ref {eq:3-1}, the projection formula is refined, and the coordinates of each point on the depth map are determined as follows: 
\begin{equation}
\begin{split}
    h=arccos(y,r)/\Delta\theta  \\
    v=arctan((z-z_r),r)/\Delta\phi
\label{eq:4-1} 
\end{split}
\end{equation}
The radius $r$ and center height $z_r$ of the reference sphere determine the extent to which the point cloud is compressed and expanded in various regions of the depth map. 

\begin{figure}[]
\centering
\includegraphics[width=12.5 cm]{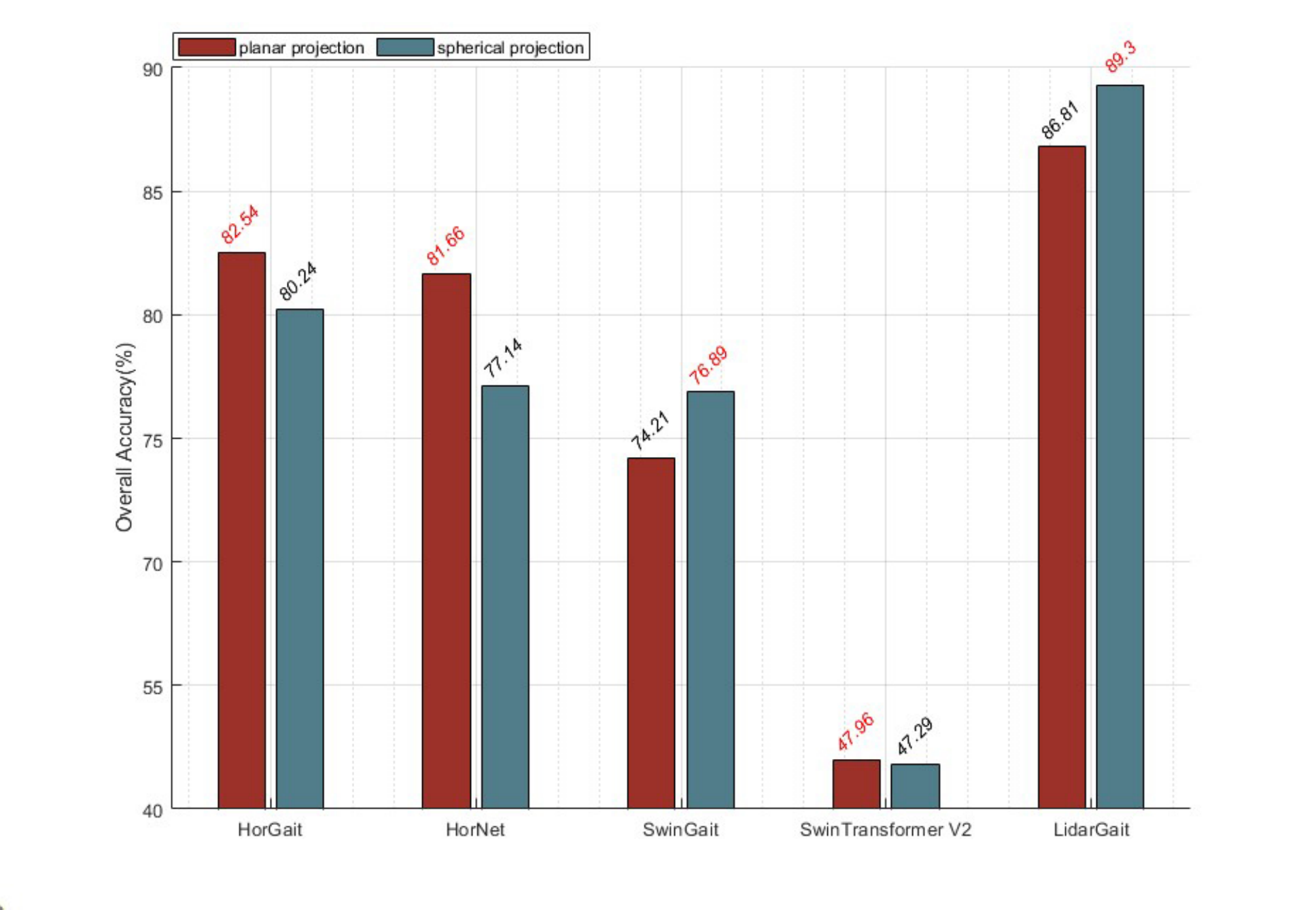}
\caption{Comparison of planar and spherical projections in gait recognition methods reveals that spherical projection excels in CNN approaches, while planar projection generally outperforms in Transformer frameworks, with the exception of SwinGait. This is due to the excessive weighting of the CNN stage in SwinGait, which undermines the advantages of the Transformer framework. }\label{fig10}
\end{figure}

Figure~\ref{fig10} demonstrates that spherical projection yields superior results in CNN methods represented by LidarGait, while planar projection achieves better accuracy in the recognition outcomes of Transformer architecture except SwinGait. 
In the CNN recognition method, the compression effect caused by planar projection results in the loss of some 3D dynamic features. 
Spherical projection, on the other hand, can free 3D points in dynamic positions, effectively enhancing the accuracy of CNN recognition. 
The Transformer architecture, however, performs better with planar projection, as its unique attention mechanism can directly focus on specific gait-related features. 
This shows that the high-order interaction of Transformer can pay more attention to dynamic features, and excessive 3D dynamic feature enhancement has an adverse effect on it.

Although SwinGait employs a Transformer structure, its performance aligns more closely with CNNs due to the extensive use of CNN stages, rendering the advantage of spherical projection less pronounced in SwinGait. 
Although SwinGait utilizes a Transformer structure, it performs better with spherical projection. 
This is because, unlike other Transformer-based methods, SwinGait incorporates CNN blocks in the first two stages, with Transformer blocks occupying only the latter two stages. 
As a result, SwinGait's Transformer architecture is incomplete, leading to insufficient attention to the planar projection of point clouds. 
Therefore, spherical projection is needed to enhance dynamic 3D features as a supplement. 
By comparison, we can conclude that HorGait fully utilizes the advantages of the Transformer structure, achieving superior gait recognition results even with limited 3D dynamic features. 
This demonstrates the strong potential of the Transformer method in gait recognition applications. 

\section{ Comparison between HorGait  and Related Works}
We compared HorGait with related work on the SUSTech1K dataset. Among them, the main forms of the dataset include silhouette, 3D point cloud, multimodal, and projection. We compared HorGait with SOTA methods for these forms. 
We also tested two Transformer methods, SwinKit and SwinTransformerV2, and a hybrid model method with Transformer architecture, HorNet, based on the projection method. 
They were compared with HorGait as SOTA methods for Transformer architecture. 
Table  ~\ref{t1} presents a comparative analysis of recognition accuracy between HorGait and related methods across a part of conditions. 
It is evident that the recognition accuracy of the hybrid model method with Transformer architecture based on projection significantly surpasses that of recognition methods relying on silhouettes and 3D point clouds. 
This indicates that projection is the optimal approach for capturing 3D features while maintaining point-to-point correlation in point clouds. 
In the comparison of Transformer architectures, HorGait exceeds all state-of-the-art methods, marking a significant advancement in the application of Transformer frameworks to gait recognition. 

Although HorGait shows slightly lower overall accuracy compared to the state-of-the-art CNN method, it delivers superior results in scenarios involving umbrellas, clothing, and carrying items, offering valuable insights for applications in specialized contexts.
While multimodal methods generally achieve higher accuracy compared to single-modal approaches, HorGait has significantly contributed to bridging the gap between these two methods. 
Although the Transformer architecture cannot surpass the above two methods in terms of recognition accuracy, it leverages its attention mechanism more effectively when datasets contain fewer 3D dynamic features. 
Its advantage in integrating prior information allows for a more comprehensive approach to recognition, and it is expected to excel in gait recognition within more complex environments in the future. 

\begin{table}[]
\caption{Evaluation with different attributes on SUSTech1K valid + test set. We compare our method with silhouette-based SOTA method GaitBase, 3D point cloud-based SOTA method PointMLP and  PointGait, SOTA Multimodal method HMRGait , SOTA Projection-based LiDARGait, SOTA Transformer method  SwinGait and SOTA hybrid model method in Normal, Clothing, Carrying, Umberalla, Night conditions. \label{t1}}

  \centering

\begin{tabular}{c|c|c|c|ccccc}
\hline
\multirow{2}{*}{Architecture} & \multirow{2}{*}{Input}                                                     & \multirow{2}{*}{Methods} & \multirow{2}{*}{\begin{tabular}[c]{@{}c@{}}Overall\\ (Rank-1 acc)\end{tabular}} & \multicolumn{5}{c}{Probe Sequence (Rank-1 acc)}                                                                                      \\ \cline{5-9} 
                              &                                                                            &                          &                                                                                 & \multicolumn{1}{c|}{Normal} & \multicolumn{1}{c|}{Clothing} & \multicolumn{1}{c|}{Carrying} & \multicolumn{1}{c|}{Umberalla} & Night \\ \hline
\multirow{5}{*}{CNN}          & Silhouette                                                                 & GaitBase\cite{opengait}& 77.50                                                                           & \multicolumn{1}{c|}{83.09}  & \multicolumn{1}{c|}{50.95}    & \multicolumn{1}{c|}{76.98}    & \multicolumn{1}{c|}{77.34}     & 26.65 \\ \cline{2-9} 
                              & \multirow{2}{*}{\begin{tabular}[c]{@{}c@{}}3D Point\\  Cloud\end{tabular}} & PointMLP\cite{pointMLP}& 68.86                                                                           & \multicolumn{1}{c|}{76.03}  & \multicolumn{1}{c|}{57.09}    & \multicolumn{1}{c|}{68.08}    & \multicolumn{1}{c|}{58.29}     & 70.75 \\ \cline{3-9} 
                              &                                                                            & PointGait\cite{Lidar3D2}& 57.60                                                                           & \multicolumn{1}{c|}{68.63}  & \multicolumn{1}{c|}{48.08}    & \multicolumn{1}{c|}{56.77}    & \multicolumn{1}{c|}{35.60}     & 61.70 \\ \cline{2-9} 
                              & Multimodal                                                                 & HMRGait\cite{Lidar3D2}& 90.23                                                                           & \multicolumn{1}{c|}{92.71}  & \multicolumn{1}{c|}{79.55}    & \multicolumn{1}{c|}{90.27}    & \multicolumn{1}{c|}{83.14}     & 90.35 \\ \cline{2-9} 
                              & \multirow{4}{*}{Project}                                                   & LiDARGait\cite{lidar1}& 86.77                                                                           & \multicolumn{1}{c|}{91.80}  & \multicolumn{1}{c|}{74.56}    & \multicolumn{1}{c|}{89.03}    & \multicolumn{1}{c|}{67.50}     & 90.41 \\ \cline{1-1} \cline{3-9} 
\multirow{3}{*}{Transformer}  &                                                                            & SwinGait\cite{deep}& 74.21                                                                           & \multicolumn{1}{c|}{77.72}  & \multicolumn{1}{c|}{64.67}    & \multicolumn{1}{c|}{74.95}    & \multicolumn{1}{c|}{55.72}     & 74.21 \\ \cline{3-9} 
                              &                                                                            & HorNet\cite{Hor}& 81.66                                                                           & \multicolumn{1}{c|}{89.80}  & \multicolumn{1}{c|}{58.06}    & \multicolumn{1}{c|}{81.49}    & \multicolumn{1}{c|}{70.54}     & 80.38 \\ \cline{3-9} 
                              &                                                                            & \textbf{HorGait}& 82.54
& \multicolumn{1}{c|}{89.45
}  & \multicolumn{1}{c|}{59.45
}    & \multicolumn{1}{c|}{83.04
}    & \multicolumn{1}{c|}{72.33
}     & 80.90
\\ \hline
\end{tabular}

\end{table}

\section{Discussion and Conclusions}

In this study, we proposed a LiDAR-based gait recognition method built on the Transformer architecture, referred to as HorGait. 
Our approach presents several significant improvements over the current SOTA Transformer models for gait recognition. 

Firstly, this marks the debut of a hybrid model combining CNN and Transformer architecture in LiDAR gait recognition. 
This architecture effectively addresses the persistent dumb patch problem in Transformer-based gait recognition, significantly enhancing its recognition accuracy in LiDAR data. 

Secondly, HorGait introduces a block go that enables adaptive input, long-range, and high-order spatial interactions, eliminating the need for CNN blocks in gait recognition stages, thereby preserving the integrity of the Transformer process. 
Additionally, in the comparison of projection methods, it is evident that the complete Transformer process, with its strong interactive capabilities, extracts more dynamic features, simplifying the data processing steps. 

Furthermore, we examined the impact of the order of high-order interactions in hybrid models. 
It was concluded that same-order combinations underperform compared to progressive order combinations, with the [1,1,3,3] configuration yielding the highest recognition accuracy. 

Looking forward, several avenues for future research could be explored. 
Firstly, the comparison with spherical projection demonstrates that the Transformer method has unique advantages that CNN cannot replicate, indicating that the application of the Transformer architecture in gait recognition holds significant potential for further improvement. 
In addition, current research on gait recognition predominantly focuses on miniaturized models, with few studies exploring large-parameter models. 
Increasing the number of network parameters could potentially overcome the limitations of the Transformer architecture and broaden its application in gait recognition. 

In summary, while our proposed HorGait model represents a notable breakthrough in LiDAR gait recognition, further refinement and optimization are required to unlock its full potential for high-precision, large-scale applications. 
Our approach not only advances the use of Transformer architecture in gait recognition but also introduces a novel approach to dynamic feature extraction, which may have a significant impact in other fields as well.

\bibliographystyle{unsrt}  


\end{document}